\documentclass[letterpaper, 10 pt, conference]{ieeeconf}  

\IEEEoverridecommandlockouts                              

\usepackage{graphicx} 
\usepackage{amsmath} 
\usepackage{amssymb}  
\usepackage{siunitx}

\title{\LARGE \bf
Universal-jointed Tendon-driven Continuum Robot: \\Design, Kinematic Modeling, and Locomotion in Narrow Tubes
}

\author{Chengnan Shentu
and Jessica Burgner-Kahrs
\thanks{
All authors are with the Continuum Robotics Laboratory, Department of Mathematical and Computational Sciences, University of Toronto, Mississauga, ON L5L~1C6, Canada 
({\tt\small cshentu@cs.toronto.edu})
}%
}

\begin{document}

\maketitle
\thispagestyle{empty}
\pagestyle{empty}




\section{Introduction}

Tendon-driven Continuum Robots (TDCRs) are promising candidates for applications in confined spaces due to
their shape, compliance, and miniaturization capability.
Non-parallel tendon routing for TDCRs have shown definite advantages including segments with higher degrees of freedom, larger workspace and higher dexterity \cite{Starke_Burgner_2017_OnTheMerits}. However, most works have focused on parallel tendons to achieve constant-curvature shapes, which yields analytically simple kinematics but overly restricts the design possibilities. 

We believe this under-utilization of general tendon routing can be attributed to the lack of a general kinematic model that estimates shape from only tendon geometry and displacements. 
Cosserat rod-based models are capable of modeling general tendon routing \cite{Rucker_Webster_2011_StaticsAndDyanmics}, but they require accurate tendon tension measurements and extensive system identification, hindering their usability for design purposes.
Recent attempts in developing a kinematic model, e.g.\cite{Bhalkikar_Ashwin_2024_KinematicModelsFor}, are limited to simple scenarios like actuation with a single tendon or tendons on perpendicular planes. 
Moreover, model formulations are often disconnected from hardware, making designs challenging to build under manufacturing constraints.

Our first contribution is a novel design for TDCRs based on a synovial universal joint module, which provides a mechanically discretized and feasible design space. Based on the design, our second contribution is the formulation and evaluation of an optimization-based kinematic model, capable of handling actuation of multiple general routed tendons.
Lastly, we present an example application of our design for gaited locomotion, demonstrating our method's potential for an unified model-based design pipeline.

\section{Mechanical Design}
\begin{figure}[t!]
  \centering
  \includegraphics[width=3.3in, height=2in]{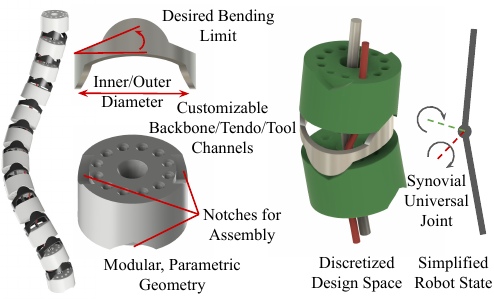}
  \vspace*{-0.5em}
  \caption{The synovial universal-jointed TDCR design restricts robot motion to achieve torsional rigidity and efficient state representation through a set of discrete joint angels. The modules' geometry is parametrically determined by tube dimensions and joint angle limits.}
  \vspace*{-1em}  
  \label{figure_design}
\end{figure}

Various discrete-jointed TDCR designs (e.g. revolute, universal, spherical) have been proposed in literature, among which universal joints show an ideal trade off between dexterity and torsional stiffness \cite{Li_Guo_2024_DesignAndKinematics}. 
However, traditional universal joints are difficult to integrate into TDCRs due to the center channel being blocked and high part count. We mitigate these major disadvantages by proposing a novel synovial universal joint with an open center-channel for backbone/tendon/tool access and simple geometry for miniaturization, shown in Fig. \ref{figure_design}. The geometry is parametrically governed by the desired inner diameter, outer diameter, and bending angle limit. It is designed to be laser-cut from a tube for low-cost production regardless of its size.

Rigid links with grooves are placed between universal joints. Their modular design allows flexibility in terms of length, diameter, and customizable tendon channels. They can be made with rapid prototyping techniques to allow iterative, application-specific designs while reusing the same set of universal joints.

\begin{figure*}[hbt!]
  \centering
  \includegraphics[width=7in, height=2in]{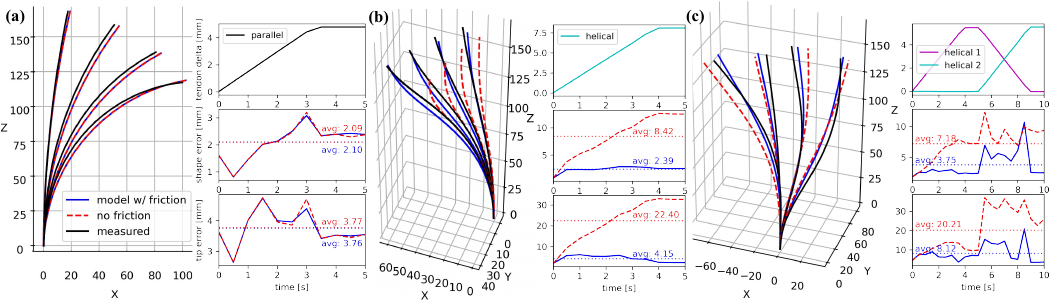}
  \vspace*{-1.5em}
  \caption{The proposed optimization-based kinematic model is evaluated on a $\SI{170}{\milli\meter}$ prototype with the actuation of \textbf{(a)} one parallel tendon, \textbf{(b)} one helical tendon, \textbf{(c)} two helical tendons simultaneously. Our model estimates shape accurately and outperforms the baseline model that omits tendon friction.}
  \vspace*{-1em}
  \label{figure_model_result}
\end{figure*}

\section{Kinematics Modeling}
An inherent benefit of our proposed design is that it simplifies the robot shape to a chain of rigid links connected by universal joints. For a robot with $n$ joints, we define the state as its joint angles $\mathbf{q}\in\mathbb{R}^{2n}$. The robot geometry is described by its link lengths \boldmath$\ell$\unboldmath \ $\in \mathbb{R}^{n}$, and relative tendon way points $\Bar{\mathbf{p}}_i \in \mathbb{R}^{3\times(2n+1)}$ for each tendon $i$ with respect to the previous joint (3D coordinates for 2 way points at each link and 1 extra way point at the base). Note this is a geometrically exact representation instead of an approximation as in other rigid-body-based model formulations.

We adopt two assumptions used in the constant-curvature model: 1) the robot's bending stiffness is symmetric and uniform, which holds for TDCRs with isotropic backbone(s) of constant cross-section; 2) we neglect external forces including gravity as they are either insignificant (e.g. low deformation due to gravity) or cannot be measured effectively so are treated as disturbances instead.

Besides that, we allow general tendon routing along the robot and assume 3) tendons have zero bending-stiffness, which holds for common tendon materials. Our state representation $\mathbf{q}$ also assumes 4) the robot is torsionally rigid, which is guaranteed by the mechanical design.

\subsection{Optimization-based Kinematics}
Having assumption 1) allows the simplification of removing material properties to formulate a \textit{dimensionless} model: Given a robot state $\mathbf{q}$, we find absolute tendon way points $\mathbf{p}_i(\Bar{\mathbf{p}}_i, \mathbf{q})$ in the robot's base frame through rigid body forward kinematics. Assuming unit tendon tension at the base, we calculate the intermediate tension vectors between all way points using the standard Capstan friction model $F_i/F
_{i-1} = e^{\gamma \mu \theta}$ where $\mu$ is the coefficient of friction, $\theta$ is the tendon wrap angle, and $\gamma=\pm1$ is the direction of tendon motion. We calculate the moments applied to the robot by taking the cross product between the vector from the closest joint and the tension difference at each way point, aggregated at each joint as $\mathbf{m} \in \mathbb{R}^{2n}$. Since the robot bends solely due to these moments, we construct a cost function that minimizes the difference between the normalized bending moments $\hat{\mathbf{m}}$ and the normalized joint angles $\hat{\mathbf{q}}$.

This cost function is passed to an optimizer with tendon displacements (equality constraints), joint angle limits (inequality constraints), and an initial guess $\mathbf{q}_0$ to estimate the robot state $\mathbf{q}^*$. We use sequential least squares programming optimizer in our current implementation. When more than one tendon is actuated, we append additional states to the optimization problems to estimate the tendon tension relative to the first tendon, which has an inequality constraint of $\geq 0$ since tendons cannot be pushed.

\subsection{Experimental Evaluation}
We evaluate our model on a $\SI{170}{\milli\meter}$ prototype consisting of parallel and helical tendons. It's actuated by servo motors (XC330-T288-T, Dynamixel, USA) with ground truth shape measured by a Fiber-Bragg Grating sensor (FBGS International NV, Belgium). A subset of shapes is used to calibrate coefficient of friction, tendon stretch, and sensor placement. The test cases and results are summarized in Fig.~\ref{figure_model_result}. The proposed model achieves accurate shape estimation, with average tip error of $\SI{8.12}{\milli\meter} \pm \SI{4.55}{\milli\meter}$ ($4.8\% \pm 2.7\%$ of length) for the test case with two simultaneously actuated helical tendons. The model outperforms the baseline model that neglects tendon friction, demonstrating the necessity to model friction for non-parallel tendons.

\section{Tendon-driven Locomotion}
\begin{figure}[hbt!]
  \centering
  \includegraphics[width=3.3in, height=1in]{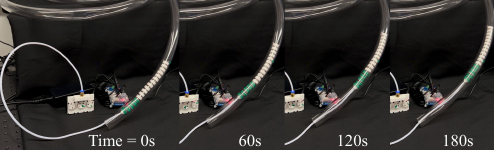}
  \vspace*{-0.5em}
  \caption{A tethered, universal-jointed TDCR climbing up a sloped rubber tube through helical rolling gait at $\SI{0.33}{\hertz}$ with three actuators. The accompanying video provides additional visual aid.}
  \vspace*{-0.5em}
  \label{figure_demo}
\end{figure}

We show an example application of our model by designing a tethered TDCR that bends into helical shapes with desired pitch angle when actuated.
This tendon configuration enables a helical rolling gait, making the prototype ($\SI{15}{\milli\meter}$ diameter) capable of locomotion in a narrow tube ($\SI{21}{\milli\meter}$ inner diameter), shown in Fig.~\ref{figure_demo}.

This tendon-driven snake-like robot forms an interesting comparison with intrinsically actuated snake robots \cite{Wright_Choset_2007_DesignOfA} --- It achieves a pre-defined locomotion gait with significantly less actuators, but loses independent joint actuation. Tendon actuation also allows such designs to be miniaturized, light-weight, low-cost, and inherently compliant, making this approach attractive for search and rescue applications.
In future works, we will evaluate the model on more complex tendon routings, and improve the computational performance of the model for real-time applications.







\bibliographystyle{ieeetr}
\bibliography{references}

\end{document}